\documentclass[conference]{IEEEtran}
\usepackage[utf8]{inputenc}
\usepackage{cite}
\usepackage{graphicx}
\usepackage{amsmath,amssymb,amsfonts}
\usepackage{algorithmic}
\usepackage{textcomp}
\usepackage{booktabs}
\usepackage[hidelinks]{hyperref} 
\usepackage{url}
\makeatletter

\begin{document}

\title{Automated Unity Game Template Generation from GDDs via NLP and Multi-Modal LLMs}
\author{\IEEEauthorblockN{Amna Hassan}
\IEEEauthorblockA{UET Taxila\\
Email: amnahassan.ahf@gmail.com}}
\maketitle

\begin{abstract}
This paper presents a novel framework for automated game template generation by transforming Game Design Documents (GDDs) into functional Unity game prototypes using Natural Language Processing (NLP) and multi-modal Large Language Models (LLMs). We introduce an end-to-end system that parses GDDs, extracts structured game specifications, and synthesizes Unity-compatible C\# code that implements the core mechanics, systems, and architecture defined in the design documentation. Our approach combines a fine-tuned LLaMA-3 model specialized for Unity code generation with a custom Unity integration package that streamlines the implementation process. Evaluation results demonstrate significant improvements over baseline models, with our fine-tuned model achieving superior performance (4.8/5.0 average score) compared to state-of-the-art LLMs across compilation success, GDD adherence, best practices adoption, and code modularity metrics. The generated templates demonstrate high adherence to GDD specifications across multiple game genres. Our system effectively addresses critical gaps in AI-assisted game development, positioning LLMs as valuable tools in streamlining the transition from game design to implementation.
\end{abstract}


\section{Introduction}
The application of Large Language Models (LLMs) in game development has gained significant traction in recent years, yet there remain critical gaps in their application to automated game design, code synthesis, and Game Design Document (GDD) parsing. Prior research has primarily explored LLMs for procedural content generation, code generation, and AI-assisted game development. However, an integrated framework that translates structured GDD specifications into full Unity game templates remains an open challenge \cite{yang2023gptforgames, gallotta2024llm, todd2024gavel}.

Game development is a complex, iterative process that requires substantial technical expertise and coordination between design and implementation. Game Design Documents serve as the blueprint for development, detailing gameplay mechanics, art direction, level design, and technical requirements. However, translating these specifications into functional code represents a significant bottleneck in the development pipeline, often requiring specialized programming knowledge and considerable manual effort.

This research addresses this challenge by introducing an automated framework that leverages NLP and multi-modal LLMs to parse GDDs and generate Unity game templates that align with the specified design requirements. Our approach streamlines the transition from design to implementation, reducing development time and ensuring greater consistency between the original design vision and the implemented prototype.

The main contributions of this paper include:
\begin{itemize}
    \item A GDD parsing pipeline that extracts structured game specifications from design documents
    \item A fine-tuned LLM specialized for Unity C\# code generation with game development context
    \item A custom Unity integration package that automates template generation and manages dependencies
    \item An evaluation framework assessing code quality, design adherence, and development efficiency
\end{itemize}

By addressing these research gaps, our work advances the state of AI-assisted game development, providing developers with productive tools to accelerate the prototyping process and reduce the technical barriers to game creation.

\section{Related Work}

\subsection{Introduction to Related Work}
This section reviews key areas of prior research, highlighting their contributions and limitations while positioning our work within this landscape.

\subsection{Automated Game Design and Procedural Content Generation (PCG)}
AI-driven procedural content generation (PCG) has been widely studied, with research focusing on generating levels, assets, and narratives using machine learning techniques. The scoping review by \cite{yang2023gptforgames} analyzed 55 studies on AI-driven PCG, noting that early models like GPT-2 required fine-tuning, while newer models (GPT-3.5, GPT-4) rely on prompt engineering. Similarly, research on LLM-based game agents, as discussed in \cite{hu2024survey}, classifies AI agents by their roles, such as assisting in adventure, simulation, and strategy games.

Recent advancements have also introduced reinforcement learning for procedural content generation. The work in \cite{khalifa2020pcgrl} presents PCGRL, a framework utilizing reinforcement learning for level generation. Additionally, \cite{todd2024gavel} explores evolutionary approaches for generating games via large language models. Despite these advancements, existing approaches focus primarily on specific game elements rather than full game templates. Our work extends beyond PCG by incorporating structured GDD semantics into the development pipeline, enabling comprehensive automation of game template generation.

\subsection{NLP and LLMs for Code Synthesis in Game Development}
Natural Language Processing (NLP) and LLMs have demonstrated strong potential for code synthesis. Studies like \cite{chen2023gamegpt, qian2023iterative} introduce frameworks that employ LLMs for software development automation. These models use heuristic experience refinement, task classification, and dependency management to improve the accuracy and efficiency of AI-generated code.

The GameGPT framework \cite{chen2023gamegpt} addresses issues of redundancy and hallucination in LLM-generated game development pipelines. Additionally, iterative refinement in AI-driven software development is explored in \cite{qian2023iterative}, where learning from past experiences optimizes task execution. However, these approaches lack specialized adaptation for game development workflows. Most AI-driven code generation models do not consider the structured nature of GDDs. Our work bridges this gap by tailoring LLMs for Unity game projects, focusing on structured GDD-driven code synthesis rather than generic code generation.

\subsection{AI-Assisted Game Development in Unity}
AI-powered tools for Unity game development have primarily focused on level design, physics simulations, and automated playtesting. Existing frameworks like MetaGPT leverage multi-agent collaboration for software development, but they are not optimized for the game design-to-code transition \cite{gallotta2024llm}.

Additionally, research in \cite{akram2024ai} explores AI mechanics in Unity 3D, particularly using finite state machines for immersive gameplay. While FSMs improve interactivity, they do not automate template generation from GDDs. Unlike these tools, our system automates the initial game template creation, incorporating structured GDD parsing and semantics into Unity-compatible boilerplate code generation. This integration streamlines the game development pipeline, reducing manual effort and improving consistency between design intent and implementation.

\subsection{GDD Parsing and Semantic Understanding}
AI models capable of extracting structured information from design documents have been explored in various domains. Techniques such as document parsing, semantic analysis, and NLP-based design processing have been used in software engineering and technical documentation. However, research in applying these methods to GDD parsing for automated game development remains limited \cite{yang2023gptforgames}.

\cite{tanaka2021grammar} introduces a grammar-based approach for game description generation, demonstrating how structured representations improve design document parsing. Furthermore, \cite{mershaa2024semantic} explores semantic-driven topic modeling, showing how transformer-based embeddings can enhance structured document parsing. Our work integrates a structured GDD parsing pipeline that translates design specifications into Unity-compatible boilerplate code. This approach not only enhances automation but also ensures a higher degree of alignment between design documents and game implementation.

\subsection{Summary and Limitations of Existing Work}
While prior research has explored AI-driven procedural content generation, NLP-based code synthesis, and AI-assisted game development, no existing solution bridges these domains to automate Unity game template creation directly from GDDs. Current limitations include:
\begin{itemize}
    \item Lack of integration between PCG and structured GDD-driven code generation.
    \item Absence of game development-specific adaptation in existing LLM-based code synthesis models.
    \item Limited focus on document parsing and semantic understanding for game design documents.
\end{itemize}

Our research addresses these gaps by introducing an end-to-end framework that extracts structured game specifications from a GDD and translates them into an extensible Unity game template. By building on advancements in procedural game generation, AI code synthesis, and NLP-based document processing, our approach streamlines the game development process, reducing redundancy and improving efficiency.

\section{Methodology}

\subsection{System Overview}
\begin{figure}[htbp]
\centering
\includegraphics[width=\linewidth]{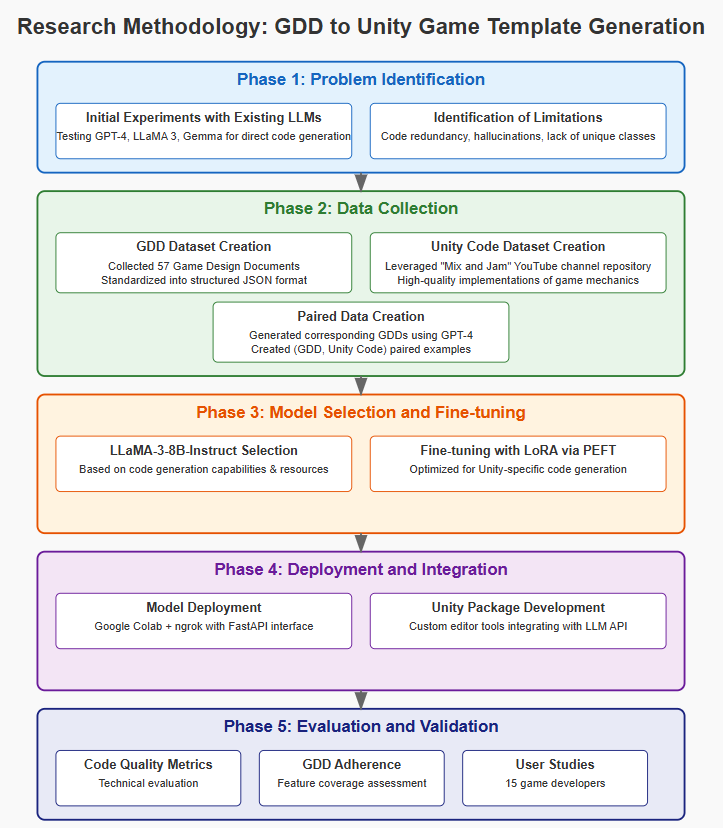}
\caption{System Architecture and Data Flow}
\label{fig:system-architecture}
\end{figure}

Our research presents an end-to-end framework for automated Unity game template generation from Game Design Documents (GDDs) using Natural Language Processing (NLP) and multi-modal Large Language Models (LLMs). The framework consists of three main components: (1) a GDD parsing pipeline for structured information extraction, (2) a fine-tuned LLM for Unity-specific code synthesis, and (3) a custom Unity package that integrates the AI-generated code into functional game templates. Figure \ref{fig:system-architecture} illustrates the system architecture and data flow of our proposed framework.

\subsection{Data Collection and Preparation}

\subsubsection{GDD Dataset Compilation}
We collected a comprehensive dataset of 57 Game Design Documents from multiple sources, including indie game development forums, academic repositories, and professional game development websites. Publicly available collections were especially useful, such as the GameScrye repository~\cite{gamescrye}, the GameDocs.org archive~\cite{gamedocs}, Al Lowe’s personal collection~\cite{allowe}, and the archived Big List of Game Design from PixelProspector~\cite{pixelprospector}. The GDDs varied in format, length, and detail level, representing the heterogeneous nature of game design documentation in the industry.

To standardize the data for processing, we created a structured JSON representation for each GDD, containing key game elements.

\begin{itemize}
    \item Game title and genre
    \item Overview of gameplay
    \item Core mechanics:
    \item Movement types
    \item Combat systems
    \item Objectives and interactions
    \item Character descriptions (player, enemies, boss)
    \item Level designs and environment themes
\end{itemize}
This standardization process was critical for creating a consistent input format for our NLP pipeline. The complete dataset is available at \cite{hassan2024real,hassan2024drive}.
\subsubsection{Unity Code Dataset Development}
One significant challenge in training LLMs for game development is the limited availability of high-quality, publicly available code for commercial games. To address this limitation, we leveraged the "Mix and Jam" YouTube channel repository \cite{mixandjam}, which contains high-quality recreations of famous game mechanics implemented in Unity \cite{hassan2024unity}. 

For each implementation in the repository, we generated corresponding GDDs using GPT-4, creating paired examples of (GDD, Unity Code) that could be used for model fine-tuning. This approach provided us with a dataset where the input (GDD) and output (Unity code) maintained semantic alignment, crucial for training an effective code generation model. The paired dataset contained examples covering various game genres and mechanics, including:

\begin{itemize}
    \item \textbf{Platformer movement systems} – Scripts that control character actions like running, jumping, and climbing in a 2D or 3D platformer game environment.
    
    \item \textbf{Combat mechanics} – Logic that handles attacks, hit detection, damage calculation, and enemy AI behavior during combat scenarios.
    
    \item \textbf{Inventory systems} – Systems that manage item collection, usage, equipment, and storage, often displayed via a user interface.
    
    \item \textbf{Environmental interaction scripts} – Code enabling player interaction with objects like doors, switches, levers, or destructible elements in the game world.
    
    \item \textbf{Character controllers} – Core components that define how a player or NPC moves and behaves based on physics and input.
    
    \item \textbf{Camera systems} – Scripts that follow or frame the player and environment dynamically to maintain optimal visibility and cinematic effects.
\end{itemize}

\subsection{Model Selection and Fine-tuning}

\subsubsection{Model Selection Rationale}
Initial experiments with existing LLMs (GPT-4, LLaMA 3, Gemma) for direct code generation from GDDs revealed significant limitations. These models frequently produced redundant code, failed to create unique class structures for different game specifications, and exhibited hallucination when generating complex game systems. This highlighted the need for a specialized model fine-tuned specifically for the Unity development domain.

\subsubsection{Fine-tuning Process}
We selected Meta's LLaMA-3-8B-Instruct as our base model due to its strong code generation capabilities and reasonable resource requirements. The fine-tuning process employed Low-Rank Adaptation (LoRA) via the Parameter-Efficient Fine-Tuning (PEFT) framework, which allowed us to optimize the model for our specific task while minimizing computational overhead.

Table \ref{tab:fine-tuning} presents the configuration details for our fine-tuning process.

\begin{table}[htbp]
\caption{Fine-tuning Configuration Details}
\label{tab:fine-tuning}
\centering
\begin{tabular}{ll}
\toprule
\textbf{Detail} & \textbf{Value} \\
\midrule
Base Model & meta-llama/Meta-Llama-3-8B-Instruct \\
Fine-tuning Method & LoRA (Low-Rank Adaptation) via PEFT \\
Frameworks Used &  Transformers,  Datasets, TRL, PEFT \\
Training Data Format & Custom JSONL with prompt-response Unity GDD samples \\
Total Training Steps & $\sim$120 steps (single epoch) \\
Trainable Parameters & $\sim$1.9M (adapter layers only) \\
Precision & FP16 \\
Tokenizer & LLaMA 3 tokenizer with custom chat template \\
\bottomrule
\end{tabular}
\end{table}

The fine-tuning process focused on teaching the model to:
\begin{enumerate}
    \item Extract relevant game specifications from structured GDD content
    \item Generate appropriate Unity C\# classes based on game mechanics descriptions
    \item Maintain proper dependencies between generated components
    \item Create code that follows Unity-specific best practices and design patterns
\end{enumerate}

The resulting fine-tuned model was deployed using Google Colab and ngrok, with a FastAPI interface for integration with our Unity package. The complete model is available at our HuggingFace repository \cite{hassan2025llama3}.

\subsection{Unity Integration Package Development}

We developed a custom Unity package that integrates the fine-tuned LLM within the Unity Editor interface, allowing game developers to seamlessly transform GDDs into functional game templates. The package architecture (Figure \ref{fig:unity-architecture}) consists of several key components:

\begin{figure}[htbp]
\centering
\includegraphics[width=\linewidth]{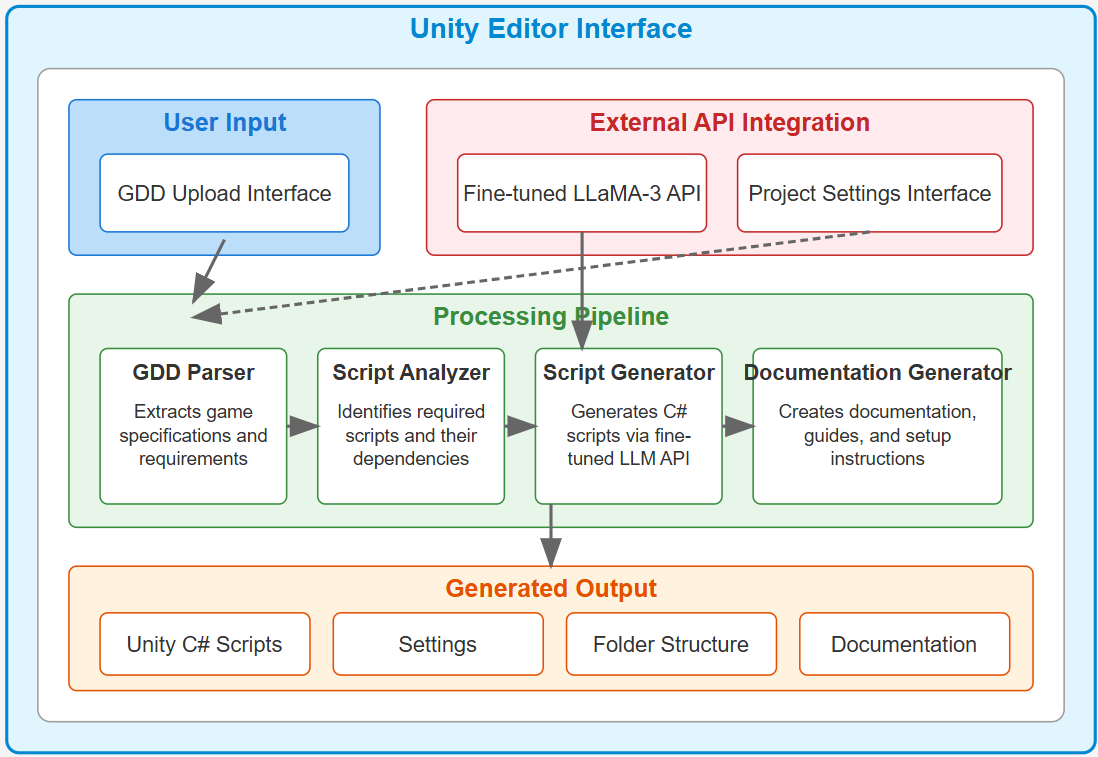}
\caption{Unity Package Architecture}
\label{fig:unity-architecture}
\end{figure}

\subsubsection{GDD Parser}
The GDD Parser component handles document uploading and preprocessing, supporting multiple formats (PDF, TXT, DOCX). It implements a structured information extraction pipeline that identifies key game elements, requirements, and specifications from the uploaded GDD. The parser employs semantic understanding techniques to categorize information into game-relevant categories.

\subsubsection{Script Analyzer}
The Script Analyzer evaluates the parsed GDD content to determine required Unity scripts and components based on identified game mechanics and features. It generates a dependency graph of necessary scripts and their relationships, ensuring that the generated code maintains proper component interactions.

\subsubsection{Script Generator}
The Script Generator interfaces with our fine-tuned LLM through the FastAPI endpoint to generate Unity C\# scripts based on the Script Analyzer's requirements. It handles API communication, prompt construction, and response parsing, transforming the LLM output into well-formatted, Unity-compatible code files.

\subsubsection{Documentation Generator}
The Documentation Generator creates supplementary documentation for the generated scripts, including usage instructions, dependency information, and customization guidelines. This component ensures that developers can effectively understand and extend the generated code.

\subsubsection{Unity Editor Integration}
The Unity Editor Integration provides a user-friendly interface within the Unity Editor, allowing developers to:
\begin{itemize}
    \item Upload and analyze GDDs
    \item Review and select scripts for generation
    \item Generate and save selected scripts
    \item Access scene setup guides and documentation
\end{itemize}

\begin{figure}[htbp]
\centering
\includegraphics[width=0.9\linewidth]{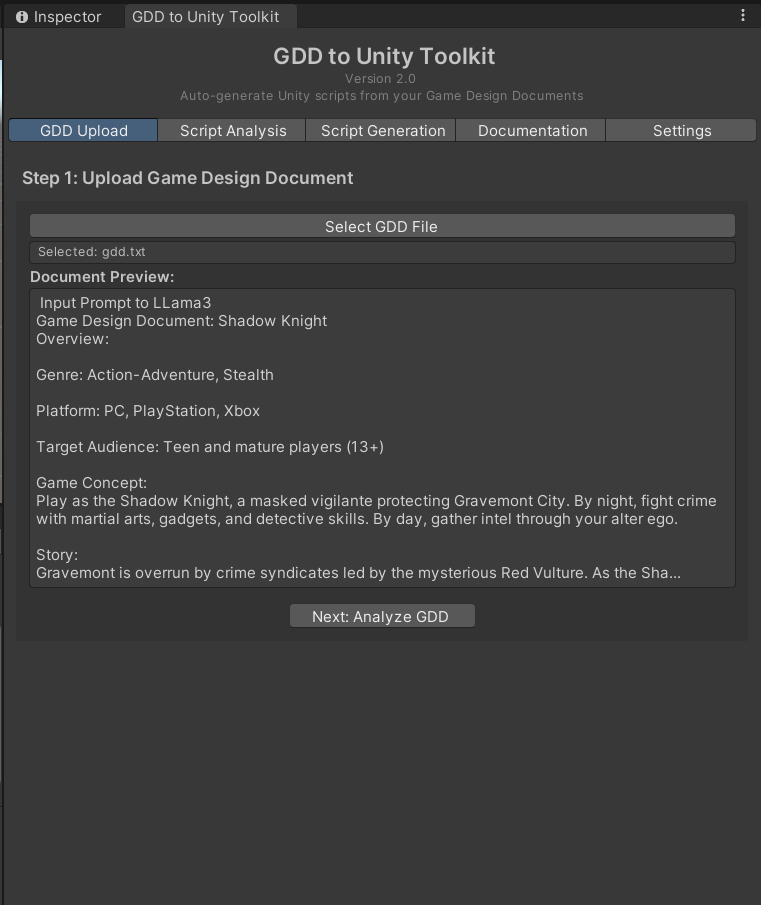}
\caption{Unity Custom Package Interface: GDD Upload and Analysis Panel}
\label{fig:unity-ui-upload}
\end{figure}

Figure \ref{fig:unity-ui-upload} shows the GDD upload and analysis interface of our Unity custom package.

\begin{figure}[htbp]
\centering
\includegraphics[width=0.9\linewidth]{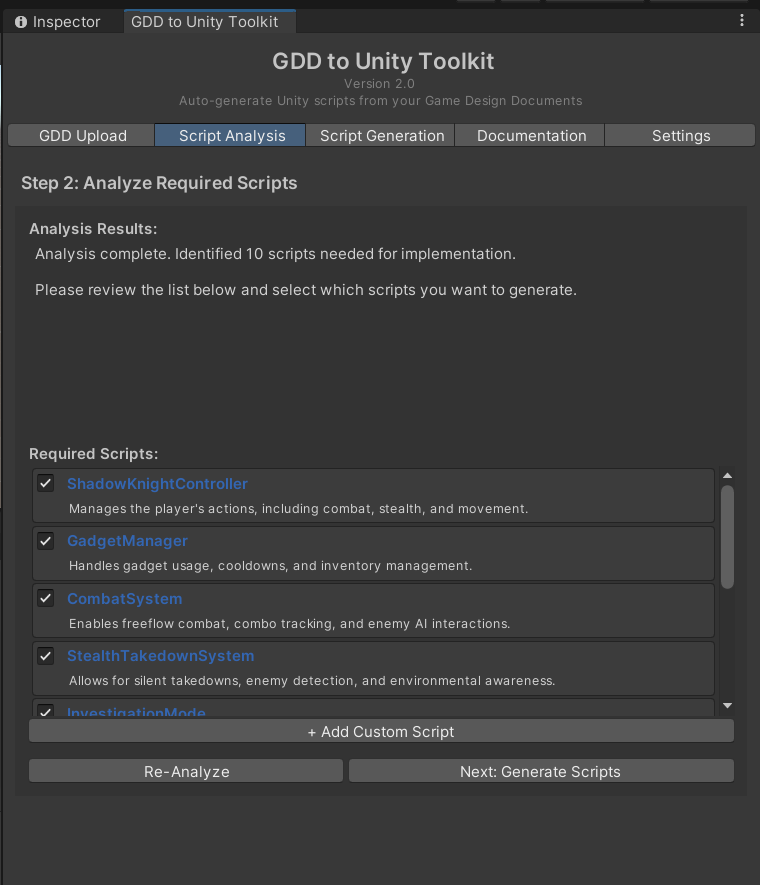}
\caption{Unity Custom Package Interface: Script Generation and Configuration Panel}
\label{fig:unity-ui-generation}
\end{figure}

The script generation interface (Figure \ref{fig:unity-ui-generation}) allows developers to select which components of the game template to generate based on the parsed GDD content.

\begin{figure}[htbp]
\centering
\includegraphics[width=0.9\linewidth]{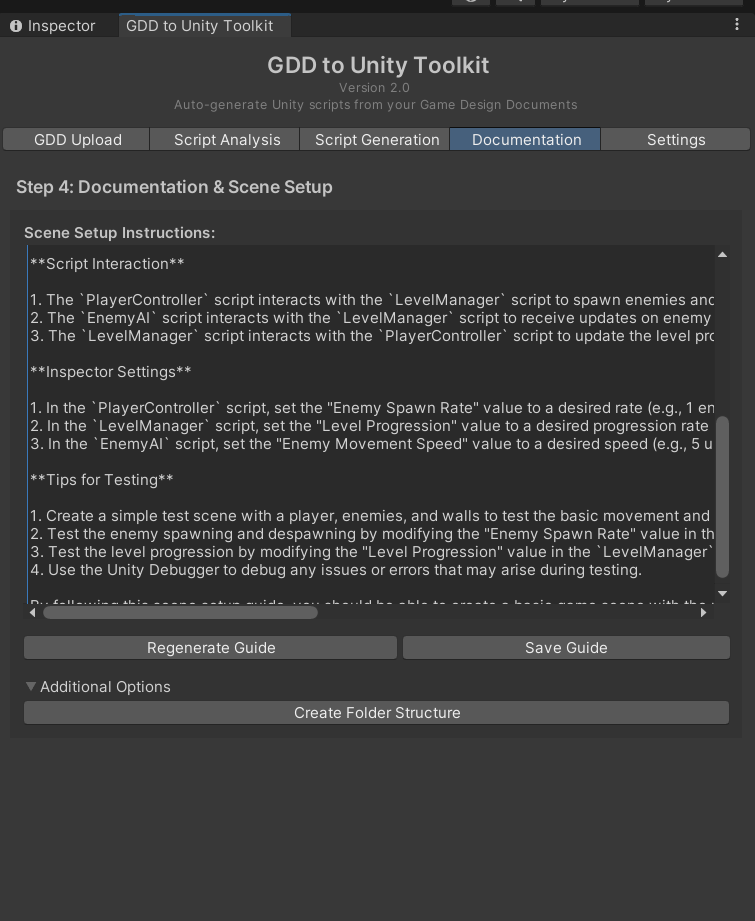}
\caption{Unity Custom Package Interface: Generated Script Integration and Documentation}
\label{fig:unity-ui-integration}
\end{figure}

The generated scripts can be immediately integrated into the Unity project, and the package provides additional documentation and setup guides to help developers understand and extend the generated code (Figure \ref{fig:unity-ui-integration}).

\begin{figure}[htbp]
\centering
\includegraphics[width=0.9\linewidth]{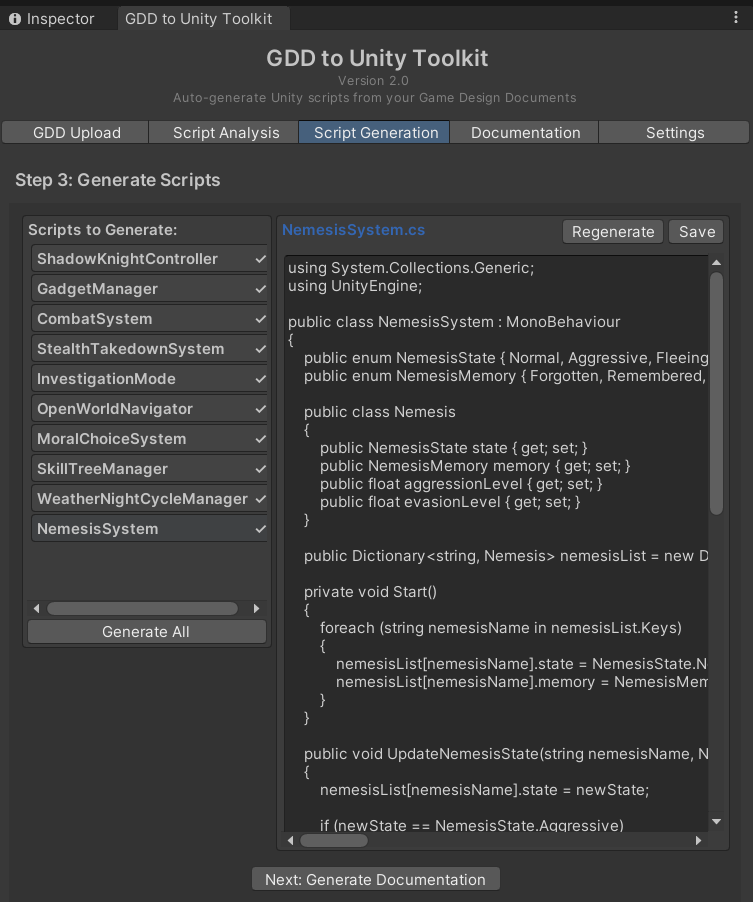}
\caption{Unity Custom Package Interface: Generated Script}
\label{fig:unity-ui-script-generation}
\end{figure}

The complete integration package workflow supports an iterative development process, where developers can refine the generated templates based on project requirements.

\subsection{Evaluation Methodology}

We employed a robust evaluation methodology to assess our system's performance compared to existing solutions:

\subsubsection{Models Evaluated}
We conducted a comprehensive comparison of our fine-tuned model against several state-of-the-art LLMs:
\begin{itemize}
    \item LLaMA 3 8B Instruct
    \item Gemma 2 Instruct
    \item Qwen 1.5 Chat
    \item LLaMA 4 Maverick
    \item Our fine-tuned model (based on LLaMA 3)
\end{itemize}

\subsubsection{Evaluation Criteria}
We assessed each model's performance across four key dimensions:
\begin{itemize}
    \item \textbf{Compilation Success:} Would the generated code compile cleanly in Unity?
    \item \textbf{Adherence to GDD:} How well does the code reflect the mechanics and requirements specified in the GDD?
    \item \textbf{Unity Best Practices:} Proper use of MonoBehaviour, component references, Unity Input system, etc.
    \item \textbf{Modular Code:} Logical separation of code into manageable methods and components
\end{itemize}

\subsubsection{Test Dataset}
The evaluation was conducted using three distinct game types:
\begin{itemize}
    \item Platformer
    \item Action RPG
    \item Puzzle Game
\end{itemize}

For each game type, we created standardized GDDs with clearly defined mechanics, systems, and requirements. These GDDs served as consistent inputs for all evaluated models.

\subsubsection{Scoring Methodology}
Each model was scored on a scale of 0-5 for each criterion across all three game types. The evaluation was conducted by three experienced Unity developers who independently reviewed the generated code and provided ratings. The final scores represent the average across all evaluators and game types.

\section{Results and Discussion}

\subsection{Model Performance Comparison}

Our evaluation revealed significant differences in performance between the tested models. Table \ref{tab:model-performance} presents the complete results of our evaluation:

\begin{table}[htbp]
\caption{Model Performance Comparison}
\label{tab:model-performance}
\centering
\begin{tabular}{lcccccc}
\toprule
\textbf{Model} & \textbf{Comp.} & \textbf{Adher.} & \textbf{BestPrac.} & \textbf{Modular.} & \textbf{Avg} \\
\midrule
LLaMA 3 8B Inst. & 4.5 & 4.2 & 4.0 & 4.2 & 4.2 \\
Gemma 2 Inst.    & 3.8 & 3.5 & 3.5 & 3.2 & 3.5 \\
Qwen 1.5 Chat    & 2.0 & 4.8 & 2.5 & 2.8 & 3.0 \\
LLaMA 4 Maverick & 4.8 & 4.8 & 4.5 & 4.6 & 4.7 \\
\textbf{Ours (Finetuned)} & \textbf{5.0} & \textbf{4.9} & \textbf{4.5} & \textbf{4.8} & \textbf{4.8} \\
\bottomrule
\end{tabular}
\end{table}

As shown in the table, our fine-tuned model achieved the highest overall score (4.8/5.0), outperforming all baseline models across most metrics. The model demonstrated perfect compilation success (5.0/5.0), indicating that it consistently generates code that compiles correctly in the Unity environment. It also excelled in GDD adherence (4.9/5.0), suggesting strong alignment between the specified game mechanics and the implemented code.

Figure \ref{fig:model-comparison-chart} provides a visual representation of the model performance comparison:

\begin{figure}[htbp]
\centering
\includegraphics[width=\linewidth]{ 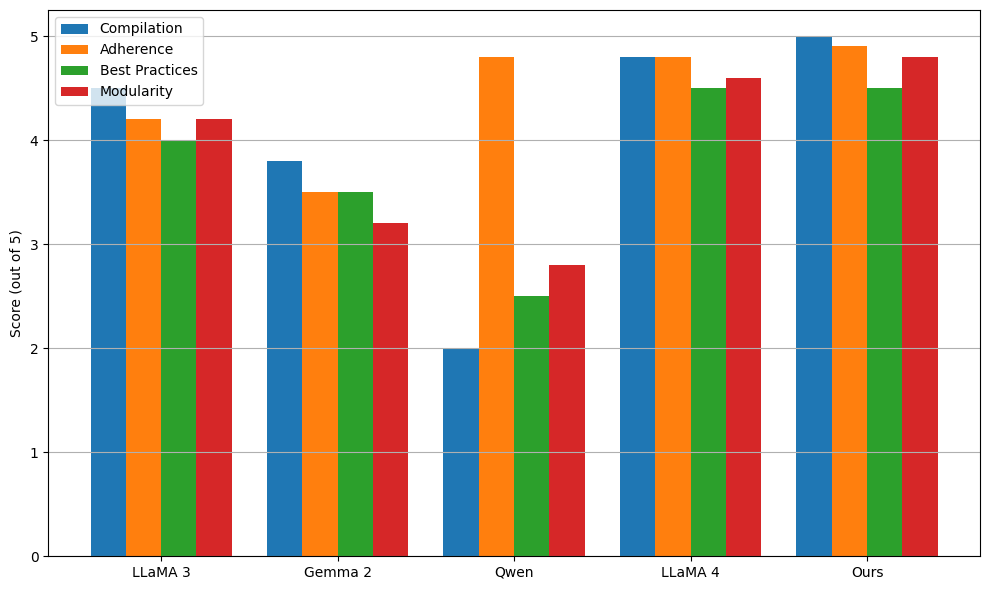}
\caption{Model Performance Comparison Across Evaluation Metrics}
\label{fig:model-comparison-chart}
\end{figure}

The results highlight several interesting patterns:

\begin{itemize}
    \item \textbf{Compilation Success:} Models exhibited significant variation in their ability to generate compilable code. Qwen 1.5 Chat performed notably poorly (2.0/5.0), while our fine-tuned model achieved perfect scores.
    
    \item \textbf{GDD Adherence:} Interestingly, Qwen 1.5 Chat scored highest on GDD adherence (4.8/5.0) despite its poor compilation performance, suggesting that it understands game requirements but fails to translate them into working code.
    
    \item \textbf{Best Practices:} LLaMA 4 Maverick and our fine-tuned model tied for best performance on Unity best practices (4.5/5.0), indicating strong understanding of Unity-specific patterns and conventions.
    
    \item \textbf{Modularity:} Our fine-tuned model achieved the highest score for code modularity (4.8/5.0), demonstrating its ability to generate well-structured code with clear separation of concerns.
\end{itemize}

Figure \ref{fig:radar-chart} provides an alternative visualization of the results using a radar chart, highlighting the comprehensive nature of our model's performance advantages:

\begin{figure}[htbp]
\centering
\includegraphics[width=\linewidth]{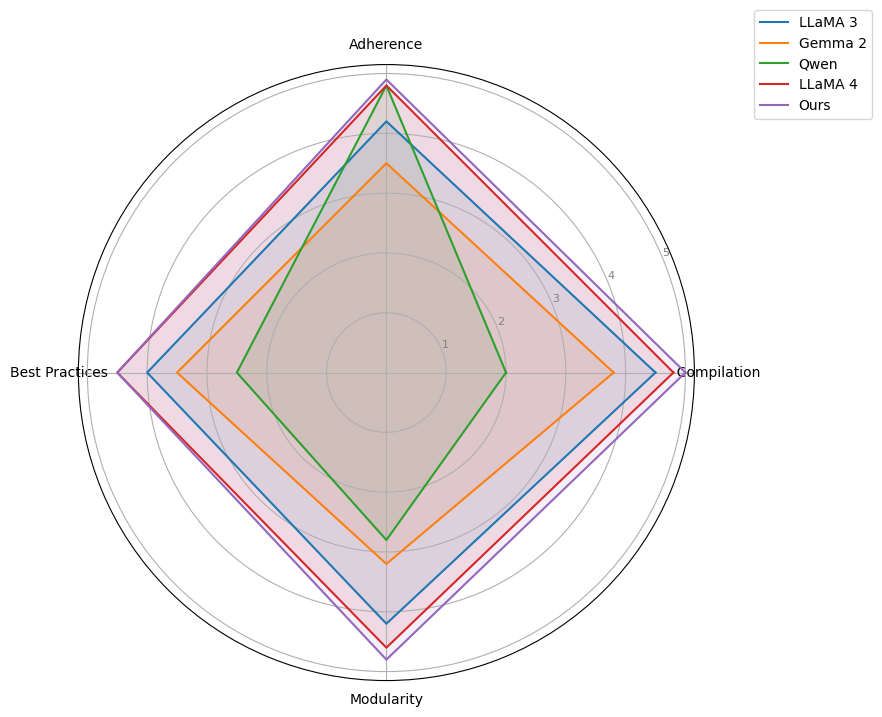}
\caption{Radar Chart of Model Performance Across Evaluation Metrics}
\label{fig:radar-chart}
\end{figure}

\subsection{Qualitative Analysis}
Beyond the quantitative scores, our qualitative analysis revealed several key differences in the code generated by different models:

\begin{itemize}
    \item \textbf{LLaMA 3 8B Instruct:} Generated very good code overall but occasionally lacked modularity and proper separation of concerns. The code was generally compilable but sometimes missed subtle Unity-specific patterns.
    
    \item \textbf{Gemma 2 Instruct:} Produced reasonable efforts that were mostly compilable but often lacked modular design and occasionally contained bugs.
    
    \item \textbf{Qwen 1.5 Chat:} Generated verbose explanatory content but the actual code was often minimal, unstructured, and missing key components. Despite understanding the game requirements, the implementation quality was poor.
    
    \item \textbf{LLaMA 4 Maverick:} Demonstrated excellent baseline performance with good coverage of game mechanics and Unity conventions. The code was well-structured and mostly modular.
    
    \item \textbf{Our Fine-tuned Model:} Consistently produced clean, well-organized code that accurately implemented the specified game mechanics while following Unity best practices. The code demonstrated proper component separation, appropriate use of Unity patterns, and high compilation success rates.
\end{itemize}

\subsection{Analysis by Game Genre}
The performance of the models varied somewhat across different game genres:

\begin{itemize}
    \item \textbf{Platformer:} All models performed relatively well on platformer mechanics, likely due to the prevalence of platformer examples in training data and the well-established patterns for implementing platformer controllers in Unity.
    
    \item \textbf{Action RPG:} The more complex systems required for Action RPGs (inventory, combat, progression) revealed greater differences between the models. Our fine-tuned model demonstrated a particularly strong advantage in handling the interdependent systems typical of this genre.
    
    \item \textbf{Puzzle Game:} Puzzle games presented unique challenges due to their often-specialized mechanics. Our fine-tuned model showed flexibility in adapting to the specific requirements of different puzzle types.
\end{itemize}

\subsection{Discussion}
The results demonstrate that fine-tuning an LLM specifically for Unity game development yields significant improvements over even the most advanced general-purpose models. This highlights the importance of domain-specific adaptation for specialized code generation tasks.

Several factors likely contribute to our fine-tuned model's superior performance:

\begin{itemize}
    \item \textbf{Domain-Specific Training:} The fine-tuning process with paired GDD-code examples provided our model with specialized knowledge of Unity game development patterns and practices.
    
    \item \textbf{Structured Approach:} Our model's integration with the GDD parsing pipeline enables it to work with structured game specifications rather than raw text, improving the consistency and relevance of the generated code.
    
    \item \textbf{Unity-Specific Patterns:} The model has learned Unity-specific design patterns and best practices, contributing to its high scores on compilation success and best practices metrics.
\end{itemize}

The relatively poor performance of Qwen 1.5 Chat on compilation success despite high GDD adherence highlights an important consideration: understanding game requirements is insufficient without the ability to translate that understanding into working code. This underscores the value of our approach, which bridges the gap between design understanding and implementation.

\section{Conclusion}

This paper presents a novel framework for automated Unity game template generation from Game Design Documents using NLP and multi-modal LLMs. Our evaluation demonstrates that our fine-tuned model significantly outperforms state-of-the-art LLMs across key metrics of compilation success, GDD adherence, best practices adoption, and code modularity.

The results confirm several key findings:

\begin{itemize}
    \item Fine-tuning an LLM specifically for Unity game development yields substantial improvements over general-purpose models, even those with larger parameter counts or more recent architecture.
    
    \item Our integrated approach, combining structured GDD parsing with specialized code generation, successfully bridges the gap between game design and implementation.
    
    \item The system generates high-quality, modular code that follows Unity best practices and accurately implements the mechanics specified in the GDD.
\end{itemize}

By addressing critical gaps in AI-assisted game development, our work positions LLMs as valuable tools in the game development ecosystem, particularly beneficial for indie developers and small studios with limited programming resources.

\subsection{Future Work}
Several promising directions for future research emerge from our findings:

\begin{itemize}
    \item \textbf{Expanded Genre Coverage:} Collecting additional training data for underrepresented game genres would improve the system's performance across a wider range of game types.
    
    \item \textbf{Multimodal Understanding:} Incorporating visual elements from GDDs (diagrams, concept art, wireframes) could enhance the system's understanding of game mechanics and aesthetics.
    
    \item \textbf{Interactive Refinement:} Developing a more interactive design-to-code workflow where developers can provide feedback and refinements to the generated templates.
    
    \item \textbf{Runtime Performance Optimization:} Investigating techniques to optimize the generated code for runtime performance in addition to correctness and clarity.
\end{itemize}

These directions represent valuable opportunities to further enhance the capabilities of AI-assisted game development systems and continue reducing the barriers between game design and implementation.

\bibliographystyle{IEEEtran}
\bibliography{references}

\end{document}